\documentclass[runningheads,a4paper]{llncs}
\usepackage{hyperref}
\usepackage{amssymb}
\usepackage{graphicx,subfigure}
\usepackage{scrextend}
\usepackage{amsmath}
\usepackage[section]{placeins}
\usepackage{gensymb}
\hypersetup{
  colorlinks   = true,
  urlcolor     = magenta,
  linkcolor    = red,
  citecolor   = blue
}

\usepackage{url}
\urldef{\mailsa}\path|{mcliche, drosenberg44, dmadeka1, cyee35}@bloomberg.net|    
\newcommand{\keywords}[1]{\par\addvspace\baselineskip
\noindent\keywordname\enspace\ignorespaces#1}

\begin{document}

\mainmatter  

\title{Scatteract: Automated extraction of data from scatter plots}

\titlerunning{Scatteract: Automated extraction of data from scatter plots}


\author{Mathieu Cliche
\and David Rosenberg\and Dhruv Madeka\and Connie Yee}
\authorrunning{Scatteract: Automated extraction of data from scatter plots}

\institute{Bloomberg LP\\
731 Lexington Ave. New York, New York, U.S.A.\\
\mailsa}

\maketitle

\begin{abstract}
Charts are an excellent way to convey patterns and trends in data, but they do
not facilitate further modeling of the data or close inspection of individual
data points. We present a fully automated system for extracting the numerical
values of data points from images of scatter plots. We use deep learning
techniques to identify the key components of the chart, and optical character
recognition together with robust regression to map from pixels to the coordinate
system of the chart. We focus on scatter plots with linear scales, which already
have several interesting challenges. Previous work has done fully automatic extraction for 
other types of charts, but to our knowledge this is the first approach that is fully
automatic for scatter plots. Our method performs well, achieving successful data extraction on
89$\%$ of the plots in our test set.
\keywords{Computer vision, Information retrieval, Data visualization}
\end{abstract}

\section{Introduction}

Charts are used in many contexts to give a visual representation of data.  However, there is often reason to extract the data represented by the chart back into a numeric form, which is easy to use for further analysis or processing.  This problem has been studied for many years, and the solution process may be divided into four steps: chart detection \cite{browuer2008segregating,ray2015architecture}, chart classification \cite{huang2003model,nair2015automated,savva2011revision}, text detection and recognition \cite{chen2015diagramflyer,mishchenko2011chart}, and data extraction \cite{mishchenko2011chart,savva2011revision}. Our work focuses on the latter two areas. \\

\begin{figure}[!htb]
\centering     
	\includegraphics[width=0.78\linewidth]{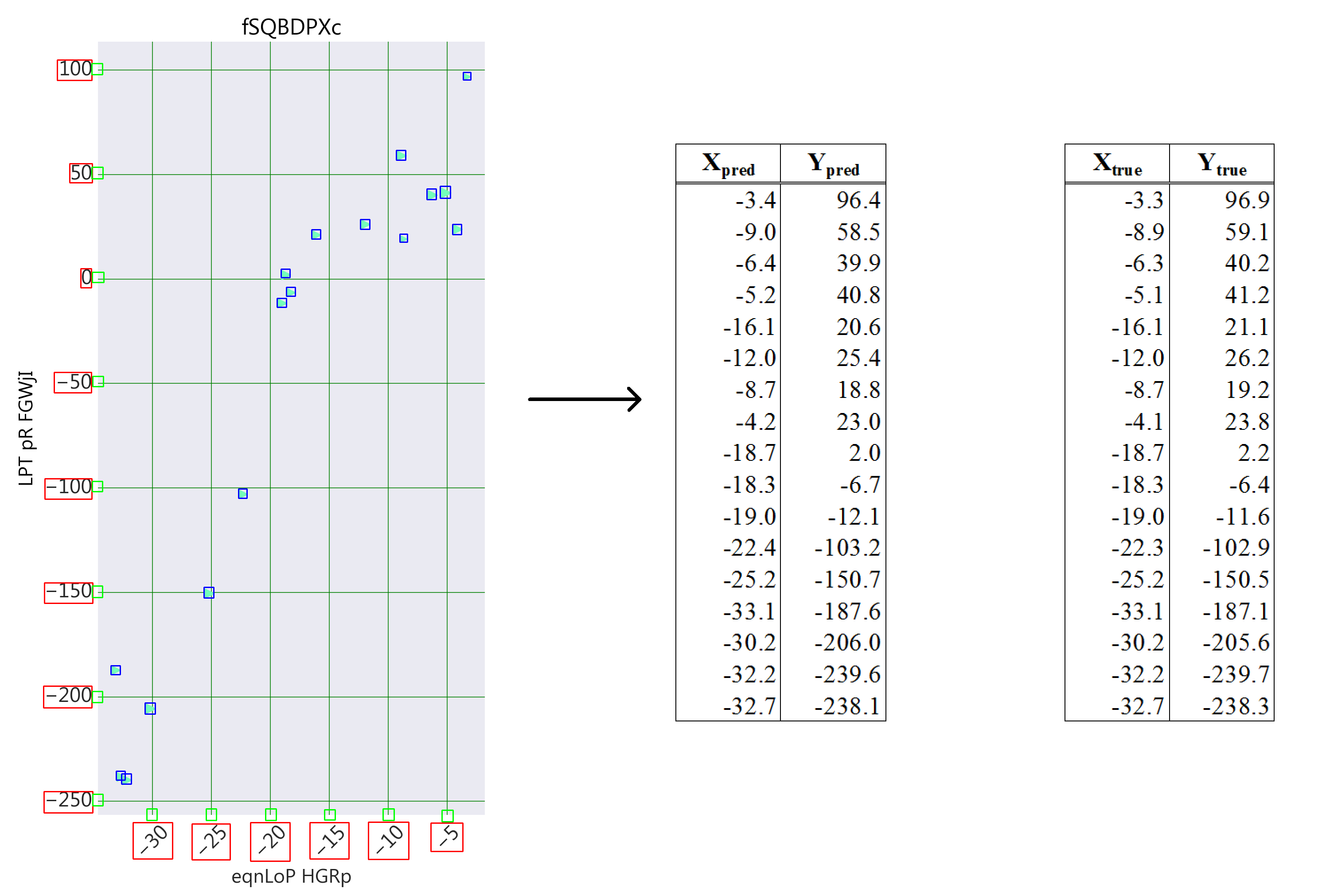}
\caption{Illustration of input and output of the the end-to-end system.  Bounding boxes are shown to illustrate the result of the object detection and are not present on the original image.  Red bounding boxes identify the tick values, green bounding boxes identify the tick marks, and blue bounding boxes identify the points.}\label{fig:system}
\end{figure}

We have built Scatteract, a system that takes the image of a scatter plot as input and produces a table of the data points represented in the plot, in the coordinate system defined by the plot axes.  There are two main steps in this process.  First, we localize three key types of objects in the image: (1) \emph{tick marks}, the visual representation of units on an axis (e.g. short lines on the axis), (2) \emph{tick values}, the representation of scale on the axis (e.g. numbers written near the tick marks), and (3) \emph{points}, the visual elements representing data tuples of the form\footnote{As is traditional, we refer to the horizontal axis as ``X'' and the vertical axis as ``Y''.} $(x, y)$.  In the chart in Fig. \ref{fig:system}, we show the bounding boxes for objects of these three types, as produced by our object detection system. The locations of these bounding boxes are initially described in \emph{pixel coordinates}, as is natural for image processing.  The second key step in our process is to map the pixel coordinates of the points into \emph{chart coordinates}, as defined by the scale on the chart axes.  The end result of our process, a table of data points, is also illustrated in Fig. \ref{fig:system}, along with the ground truth $(x,y)$ coordinates of the data.
 
While information extraction from charts and other infographics has been widely studied, to our knowledge, our system is the first to extract data from an image of a scatter plot and represent it in the coordinate system of the chart, fully automatically. The closest in aim is \cite{baucom2013scatterscanner}, an unpublished work that describes an approach to such a system, but only gives test results for the point detection part. Semi-automatic software tools are available, but they require the user to manually define the coordinate system of the chart and click on the data points, or to provide metadata about the axes and data \cite{chartsense-interactive-data-extraction-chart-images,mishchenko2011chart,shadish2009using,yang2006semi}.\\ 

There exist several systems that attempt to extract key components of charts, but do not attempt to convert from pixel coordinates to chart coordinates.  Many use heuristics, such as connected component analysis, edge detection and k-median filtering \cite{huang2007system,kataria2008automatic,lu2009automated,nair2015automated}.
Recent work has taken a machine learning approach.  \cite{al2017machine} classifies the role of extracted chart components (e.g. bar or legend).  In \cite{tsutsui2017data}, a deep learning object detection model is trained to detect sub-figures in compound figures.  FigureSeer uses a convolutional neural network to extract features for the localization of lines and heuristics for the localization of tick values \cite{siegel2016figureseer}.  The end result used for evaluation is the localization of all the relevant chart components, but does not include the pixel-to-chart coordinate transformation. \\

For charts other than scatter plots, there are two systems that aim to convert from pixel coordinates to chart coordinates.  The automatic data extraction in \cite{al2015automatic} is for bar charts.  It extracts the text and numerical values using OCR, and then recovers the numerical values for each bar by multiplying its height in pixels by the pixel-per-data ratio.  Inaccuracies of the OCR tool resulted in a significant number of the charts having incorrectly calculated Y-axis scale values.  The ReVision system proposed in \cite{savva2011revision} recovers the raw data encoded in bar and pie charts by extracting the labels with OCR and using a scaling factor to map from image space to data space.  All these systems are highly dependent on the accuracy of the OCR.\\

In this paper, we propose Scatteract, an algorithm that leverages deep learning techniques and OCR to retrieve the chart coordinates of the data points in scatter plots.  Scatteract has three key advantages over the systems mentioned above. First, to our knowledge, Scatteract is the only system to use deep learning methods to identify all the key components of a scatter plot. The second novel aspect of our work is that we created a system for procedurally generating a large training dataset of scatter plots, for which the ground truth is known, without requiring any manual labeling.  This allows our system to be much more extensible than heuristic methods built around a set of assumptions.  The plot distribution can be easily modified to accommodate new plot aesthetics, while heuristic methods may need a complete redesign to accommodate these changes.  The third key advantage of our system is that our method for determining the mapping from pixel coordinates to chart coordinates is fairly robust to OCR errors.  \\

This paper is organized as follows.  In Sec. \ref{dataset} we describe the datasets used to train and test Scatteract.  In Sec. \ref{method} we expand on the methodology and present test results for the building blocks of Scatteract.  In Sec. \ref{results} we present a performance analysis of the end-to-end system, and we outline our main conclusions in Sec.~\ref{conclusion}.

\section{Datasets}\label{dataset}

Scatteract uses an object detection model, which requires a large amount of annotated data for training.  One way to achieve this is to collect a large sample of scatter plots from the web and manually label the bounding boxes for the objects of interest (points, tick marks, and tick values).  A  more efficient approach is to generate the scatter plots procedurally, so that the bounding boxes are known.  Using this approach, we generated 25,000 train images and 600 test images. Some examples are shown in Fig. \ref{fig:suc_gen} and Fig. \ref{fig:fail_gen}.  Besides these artificial charts, we scraped an additional 50 scatter plots from the web (Fig. \ref{fig:suc_web} and Fig. \ref{fig:fail_web}).  More details on how our datasets were obtained are below.

\subsection{Procedurally generated scatter plots}

To achieve randomness in the scatter plots, we developed a script to randomly select values that affect the aesthetics and the data of the scatter plot.  We used the Python library Matplotlib \cite{Hunter:2007} to generate the plots since it allows for easy extraction of the bounding boxes of the points, tick marks, and tick values. The factors we used to randomize the plot aesthetics and data are listed below. \\
{\scriptsize
\begin{enumerate}
\item Plot aesthetics
\begin{enumerate}
\item Plot style (default styling values e.g. ``classic'', ``seaborn'', ``ggplot'', etc.)
\item Size and resolution of the plot
\item Number of style variations on the points
\item Point styles (markers, sizes, colors)
\item Padding between axis and tick values, axis and axis-labels, and plot area and plot title
\item Axis and ticks locations
\item Tick sizes if using non-default ticks
\item Rotation angle of the tick values, if any
\item Presence and style of minor ticks
\item Presence and style of grid lines
\item Font and size of the tick values
\item Fonts, sizes and contents of the axis-labels and plot title
\item Colors of the plot area and background area
\end{enumerate}

\vspace{0.25cm}
\item Plot data
\begin{enumerate}
\item Data points distribution (uniformly random, or random around a linear or quadratic distribution)
\item Number of points
\item Order of magnitude of the X and Y coordinates
\item X and Y coordinates ranges around the selected order of magnitude
\item Actual values of the X and Y coordinates given the order of magnitude, ranges and distribution
\end{enumerate}
\end{enumerate}
}
These parameters allow us to build a wide variety of scatter plots, for which we have full ground-truth labels.  Some of the plots generated from this procedure are very difficult and sometimes impossible to read, even for a human. For example, the tick values can overlap with each other or with the data points, and the randomly selected font for the tick values can be unreadable.  Although we did not eliminate such unrealistic plots from the training set, we did manually remove them from the test set.  \\

\subsection{Scatter plots from the web}

To see how our system, trained on randomly generated scatter plots, generalizes to real charts, we also collected a small test set of scatter plots that were generated by humans.  We downloaded 50 scatter plots from a Google image search for ``scatter plot''.  The only inclusion criteria was the presence of tick values without units\footnote{It is possible for our system to take units into account, but for simplicity we postpone this extension to future work.}.

\section{Methodology}\label{method}

We take as input the image of a scatter plot, and the output is a set of the ($X_{\text{pred}},Y_{\text{pred}}$) chart coordinates of the data points detected.  The pipeline is as follows: 
{\small
\begin{enumerate}
\item Use the object detection model to find bounding boxes for the tick marks, tick values, and points.
\item Apply OCR on the images inside the bounding boxes of the tick values.  
\item Find the closest tick mark to each tick value.
\item Use clustering to assign each (tick mark, tick value) pair to either the X or Y axis.
\item Apply robust regression to determine the mapping from pixel coordinates to chart coordinates, for each axis.
\item Apply the mapping to the pixel locations of the detected points to build the table of chart coordinates.  
\end{enumerate}
}
Below we expand on the most important steps.

\subsection{Object detection}

Object detection is the task of putting bounding boxes around objects that appear in an image.  For our task, we use object detection to localize points, tick marks, and tick values. There is a wide variety of object detection models, but all the state-of-the-art methods use deep learning. We chose ReInspect \cite{stewart2016end} as our object detection method because it is very effective in crowded scenes, and the points in scatter plots are often very close together, even overlapping.  ReInspect uses the OverFeat algorithm \cite{sermanet2013overfeat} but adds a long short-term memory network at the end of it, such that predictions are not made independently of each other.  We used the Tensorflow implementation of ReInspect, TensorBox\footnote{https://github.com/TensorBox/TensorBox}. It is standard practice to initialize the core of the model with a pretrained convolutional network, for which we selected Google Inception V1 \cite{szegedy2015going}.  We trained three separate models: one for tick marks, one for tick values, and one for points. We resized our scatter plot images to 800x800 pixels for the tick mark and tick value models, and to 1440x1440 for the point model to help resolve a high density of points.  Each model was trained for 5 epochs on a Geforce GTX Titan X GPU, and the total training time was about 60 hours.  Given an image, a model outputs a set of bounding boxes for the object it is detecting, along with a confidence score for each.  We only keep detections with a confidence above 0.3, based on validation experiments.\\ 

\begin{figure}[!htb]
\centering     
	\includegraphics[width=0.57\linewidth]{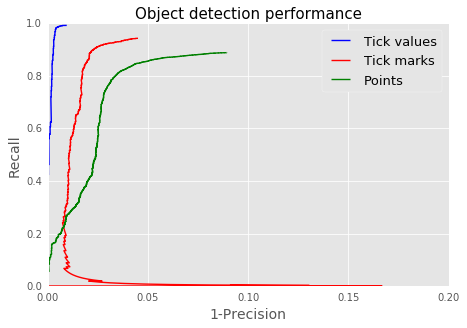}
\caption{Recall-Precision curves for the object detection.}\label{fig:pr-curves}
\end{figure}

We can evaluate the performance of the object detection on the procedurally generated test set.  We define a true positive as a predicted bounding box which has an intersection-over-union ratio with a true bounding box above 50\%.  We can then vary the threshold on the confidence of the predicted bounding boxes to generate recall-precision curves, see Fig. \ref{fig:pr-curves}.  We get an average precision of 87.1\% for the detection of points, 99.1\% for the detection of tick values and 93.0\% for the detection of tick marks.  The detection of points is the most difficult, particularly for plots with many overlapping points.  However, in the end-to-end system, failures on the detection of tick marks and tick values can be much more costly, since they are essential for determining the pixel-to-chart coordinate conversion.  \\

\subsection{Optical character recognition}

\begin{figure}[!htb]
\centering     
\subfigure[Illustration of the results from the rotation-fixing heuristics.  The blue bounding box represents the minimum area rectangle that encloses the thresholded image.]{\label{fig:ocr_rot}	\includegraphics[width=0.48\linewidth]{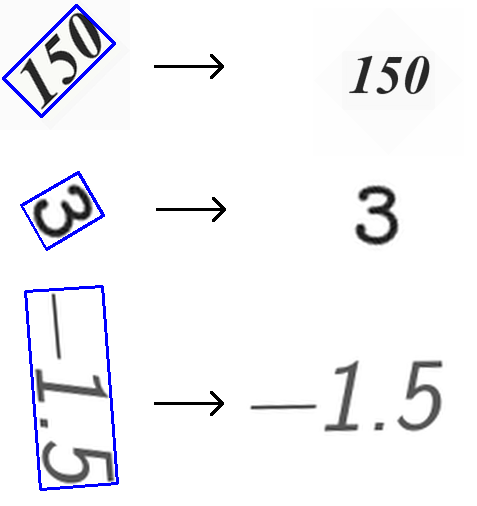}}
\hfill
\subfigure[Illustration of the clustering problem used to assign the tick marks to either the X- or Y-axis.]{\label{fig:axis_split}\includegraphics[width=0.48\linewidth]{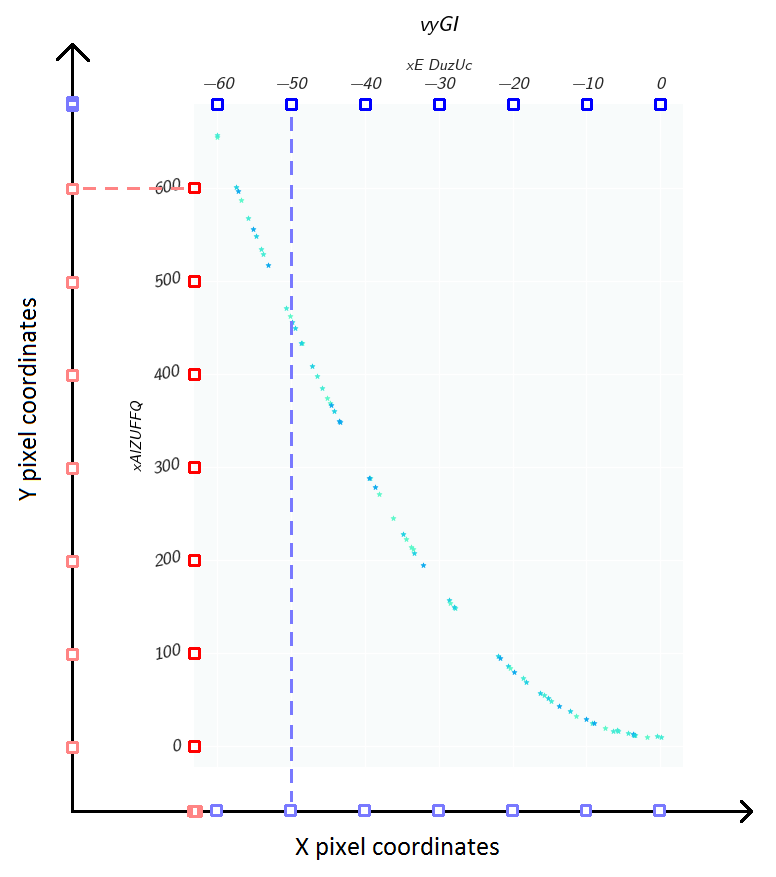}}
\caption{}
\end{figure}

To extract the tick values, we apply OCR to the content of the bounding boxes produced by the tick-value model.  We use Google's open source OCR engine, Tesseract \cite{smith2007overview}, because of its ease of use.  It comes pretrained on several fonts, and while it is possible to retrain it on custom fonts, we used it as is. However, we found that performing several preprocessing steps on a tick-value image before applying Tesseract significantly improves OCR accuracy.  We first convert the image to gray scale and then rescale it so that its height is 130 pixels, while maintaining its aspect ratio.  The most important transformation, however, is a heuristic procedure to rotate the tick-value images such that they are horizontally aligned, since this is what Tesseract is expecting.  This procedure\footnote{The procedure is largely inspired by this blog post: \\ http://felix.abecassis.me/2011/10/opencv-rotation-deskewing/.} finds the minimum area rectangle that encloses the thresholded tick-value image, and then rotates the image by an angle that takes into account the angle of the rectangle and the number of characters in the tick-value image.  The result of this procedure is illustrated in Fig \ref{fig:ocr_rot}.  It works well except for tick values that are rotated by precisely $\pm 90^{\circ}$   or upside-down.\\

To test the OCR piece, we used true tick-value images along with their corresponding tick values, for each of the plots in the procedurally generated test set.  Then, for each tick-value image, we ran our image preprocessing technique, followed by the Tesseract engine, and compared the predicted tick values with the ground truth.  This gives an accuracy of 91.2\%, but without the image preprocessing the accuracy drops to 63.8\%.

\subsection{Axis splitting}

Now that we have extracted the tick value associated with each tick-value image, we need to associate each value with its corresponding tick mark and assign it to the X-axis or Y-axis.  We make use of the observation that if we project the centers of the tick-mark bounding boxes to the X-axis, the projection of the ticks from the Y-axis are clustered in one area, while the projection of ticks from the X-axis are spread out, as illustrated in Fig. \ref{fig:axis_split}. This therefore effectively becomes a clustering problem. 
The algorithm is as follows:
\begin{enumerate}
\item Assign each tick value to its closest tick mark.  
\item Perform a 1-dimensional DBSCAN \cite{ester1996density} clustering algorithm twice: once to obtain a set of clusters for the X pixel coordinates of the ticks, and another one to obtain a set of clusters for the Y pixel coordinates of the ticks.  The expected result from the two DBSCANs is one cluster with the X coordinates,  these are the ticks that belong on the Y-axis, and one cluster with the Y coordinates, these are the ticks that belong on the X-axis.
\end{enumerate}

\begin{figure}[!htb]
\centering     
\subfigure[]{\label{fig:ransac_chart}\raisebox{12mm}{\includegraphics[width=0.43\linewidth]{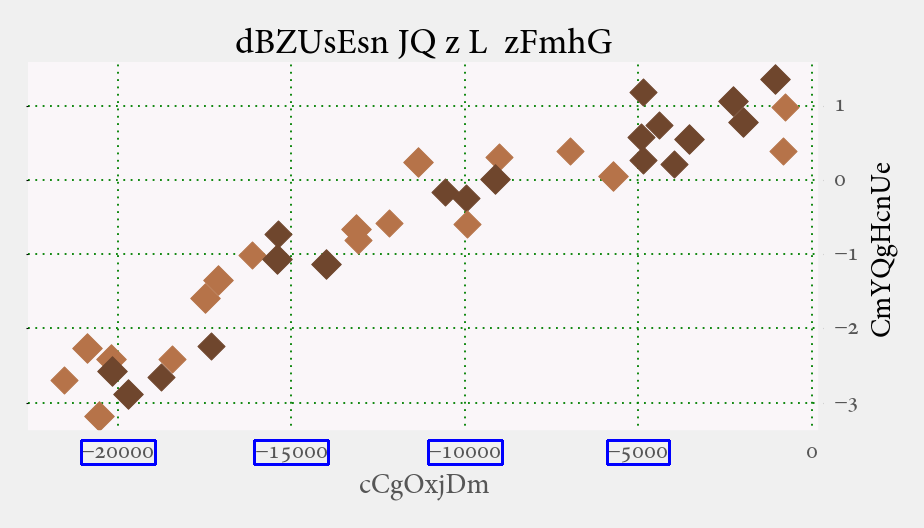}}}
\subfigure[]{\label{fig:ransac_reg_img}\includegraphics[width=0.56\linewidth]{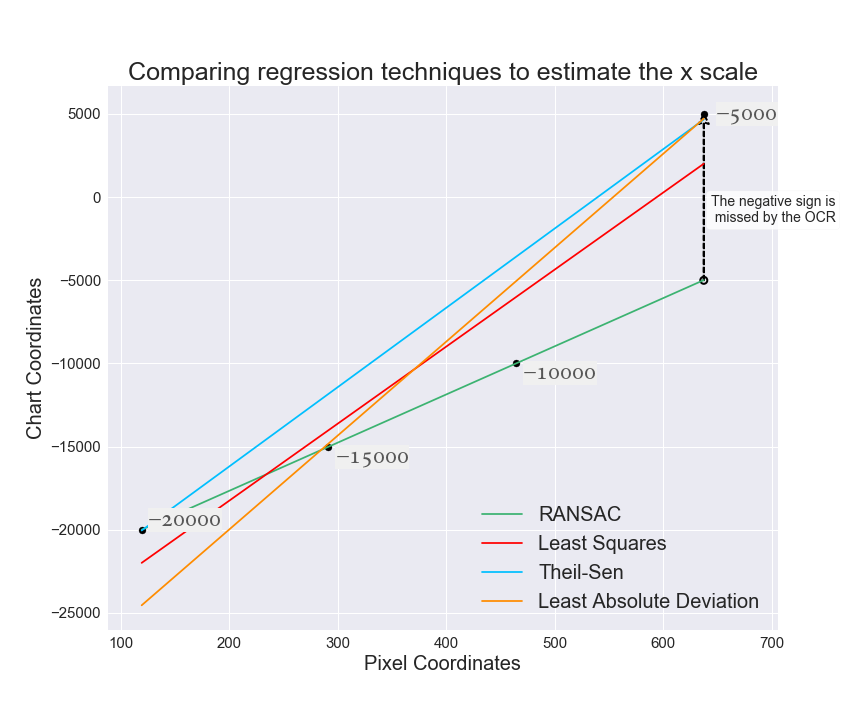}}
\caption{In Fig. \ref{fig:ransac_chart} above, the OCR fails to detect the minus sign in front of the number -5000. While other regression techniques are thrown off by this outlier, Fig. \ref{fig:ransac_reg_img} shows that RANSAC regression (visualized by the green line) provides a more robust estimation of the transformation from pixel to chart coordinates.}
\label{fig:ransac}
\end{figure}

\subsection{RANSAC regression}
At this point, we need to find the mapping from pixel to chart coordinates.  We assume the scales are linear, so the conversion is an affine transformation.  For the X-axis, it takes the form $X_{\text{chart}} = \alpha_x X_{\text{pixel}} + \beta_x$, and similarly for the Y-axis.  We now create a pair of the form $(X_{\text{pixel}} , X_{\text{chart}})$ for each tick mark\footnote{We use the center of the tick-mark bounding box as the tick mark location.}/tick value pair we find on the X-axis, and similarly for the Y-axis.  For the chart in Fig. \ref{fig:ransac_chart}, we have extracted these pairs for the X-axis and plotted them as black points in Fig. \ref{fig:ransac_reg_img}. In principle, we can determine our affine transformation just by performing linear regression on these points.  However, standard least-squares regression is very sensitive to outliers, which means that a single OCR error can cause a complete failure to estimate the transformation.  We therefore explored alternatives that are more robust to outliers, such as least absolute deviations (LAD) regression, Theil-Sen Regression, and RANSAC regression.  While all three fared better than linear regression, we obtained the best results with RANSAC regression \cite{fischler1981random}.  RANSAC regression is an iterative algorithm in which linear regression is applied to samples of the data points in order to detect inliers and outliers from the out-of-sample data points.  The model with the most inliers and the smallest residuals on the inliers is chosen.  For RANSAC we need to choose a loss function, as well as a maximum residual $R_{\max}$ beyond which a point is classified as an outlier. We use a square loss function, and choose $R_{\max}$ as the median absolute deviation of the tick values squared and divided by 50, such that for the X-axis:
\begin{eqnarray}
R_{\max,x} = \frac{\text{median}\left( X_{\text{chart},j} - \text{median}(X_{\text{chart}})\right)^2}{50}. 
\end{eqnarray}
We apply RANSAC regression for both the X- and Y-axes, such that we end up with a set of 4 parameters ($\alpha_x,\alpha_y,\beta_x,\beta_y$) for each scatter plot.  It is then straightforward to apply these affine transformations to the pixel coordinates of the center of the bounding box of each point detected, and thus obtain their location in chart coordinates. Fig. \ref{fig:ransac} shows how RANSAC improves the estimate of the affine transformation compared to other regression techniques.

\section{Results}\label{results}

\subsection{Performance analysis}

\begin{figure}[t]
\centering     
	\includegraphics[width=0.57\linewidth]{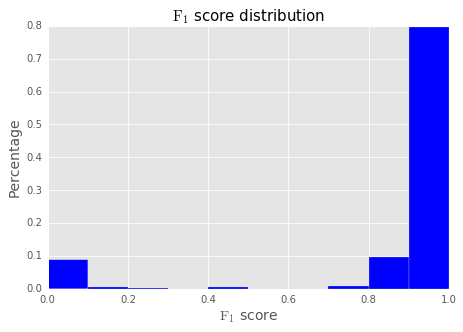}
\caption{$F_1$ score distribution across the plots in the procedurally generated test set.}\label{fig:f1_dist}
\end{figure}

To evaluate the performance of the end-to-end system, we need to have an evaluation metric.  We define true positives as predicted points that are within 2\% of true points in chart coordinates:
\begin{eqnarray}
\frac{|X_{\text{pred}} - X_{\text{true}}|}{\Delta X_{\text{true}}} <= 0.02  \quad\text{and}\quad \frac{|Y_{\text{pred}} - Y_{\text{true}}|}{\Delta Y_{\text{true}}} <= 0.02
\end{eqnarray}
where $\Delta X_{\text{true}}=\max_j(X_{\text{true},j})-\min_j(X_{\text{true},j})$ and $\Delta Y_{\text{true}}=\max_j(Y_{\text{true},j})-\min_j(Y_{\text{true},j})$ are the true ranges of values for the X and Y chart coordinates.  We compute all pairs of distances between predicted and true points, and we first select the predicted points closest to each true point in order to find true positives.  If a pair of nearby predicted and true points satisfy the true positive criteria, both true and predicted points are removed from the true and predicted set such that we do not overcount true positives. We repeat this process until either no points are remaining or none of the closest points satisfy the true positive criteria.  With this definition we can evaluate the precision and recall for each plot.  Taking the average across all plots in the procedurally generated test set, we end up with an average precision of 88$\%$ and an average recall of 87$\%$.  We can also compute the $F_1$ score for each plot. Fig. \ref{fig:f1_dist} shows the distribution of $F_1$ scores across all plots in the procedurally generated test set.  Based on Fig. \ref{fig:f1_dist}, we can see that plots either have a very high $F_1$ score or a very low one.  The latter case indicates that the pixel/chart coordinate transformation was not determined correctly.   We can therefore define a successful data extraction for a plot if the $F_1$ score is above 80$\%$ for that plot.  By this definition, our method is successful for 89$\%$ of the plots in the procedurally generated test.  We can also use this definition to evaluate alternative versions of our method.   In Table \ref{tab:cv}, note that the version that uses linear regression on 2 tick values instead of RANSAC is analogous to what \cite{baucom2013scatterscanner} used to find the conversion factors between pixel and chart coordinates.  RANSAC has a net advantage over the other methods, giving an absolute performance boost of 5.4\% over Theil-Sen, the next best method. We note that in its current implementation, Scatteract requires 3.5 seconds to extract the data from a single plot.  \\

\begin{table*}[t]
 \centering
\begin{tabular}{|c|c|}
  \hline
  System & Success rate \\
  \hline \hline
  Scatteract & \textbf{89.2$\%$} \\
  \hline
  Scatteract without image preprocessing before OCR & 43.3$\%$ \\
  \hline
  Scatteract with Theil-Sen instead of RANSAC & 83.8$\%$ \\
  \hline
  Scatteract with LAD instead of RANSAC & 82.8$\%$ \\
  \hline
  Scatteract with linear regression on 2 tick values instead of RANSAC  & 70.7$\%$ \\
  \hline
\end{tabular}
  \caption{Test results on variations of our system. }
  \label{tab:cv}
\end{table*}

Without the ground truth available for the test set from the web, we cannot evaluate it as systematically as the procedurally generated one.  However, a visual inspection of the plots is sufficient to determine if the data extraction is successful.  To quickly see if the axes were correctly decoded, we can look at a handful of predicted points and see if they correspond to actual points on the plots.  Moreover, to see if most points were detected correctly, we can inspect the predicted bounding boxes for the points directly on the plot image.  Through visual inspection we conclude that our method gave a successful data extraction on 39 plots out of the 50, or 78$\%$.  As one might expect, this is lower than on the procedurally generated test set due to the fact that these plots occasionally exhibit features which are not present in the training data.  These aspects make identifying all the relevant objects more challenging for the object detection. 

\subsection{Error analysis}

\begin{figure}[t]
  \centering
  \begin{minipage}[b]{0.48\textwidth}
	\includegraphics[width=1.0\linewidth]{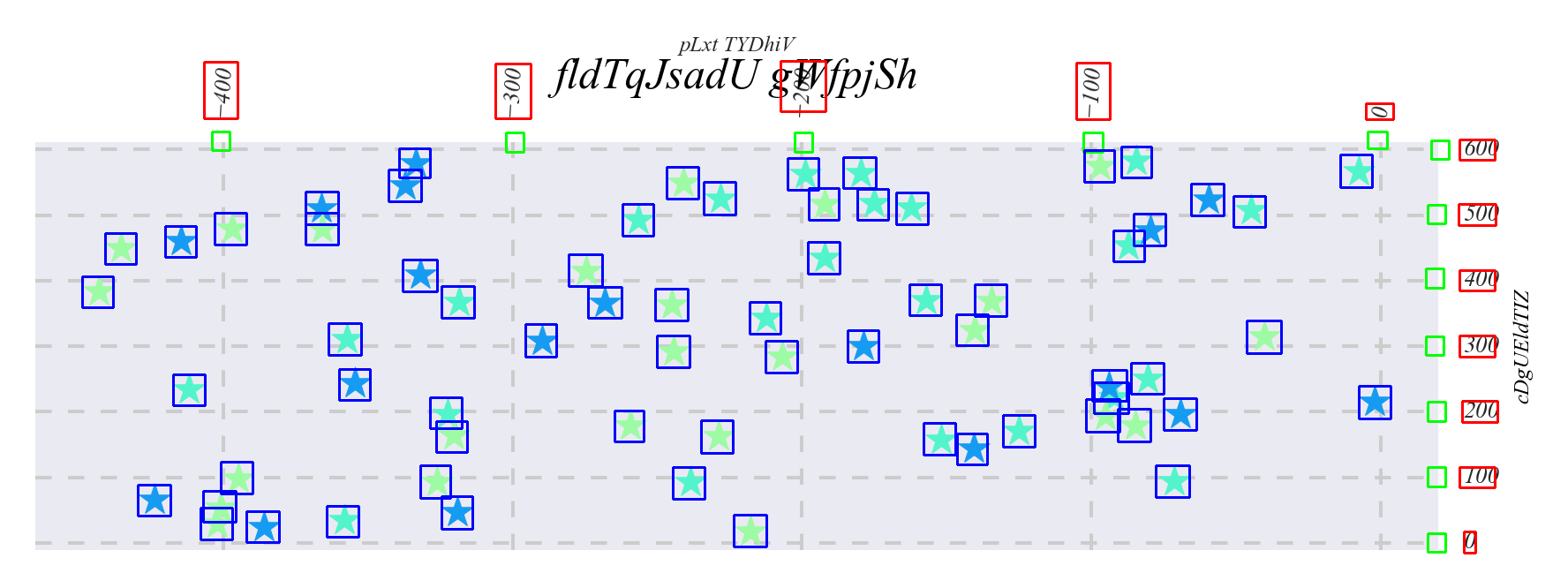}
  \end{minipage}
  \hfill
  \begin{minipage}[b]{0.48\textwidth}
	\includegraphics[width=1.0\linewidth]{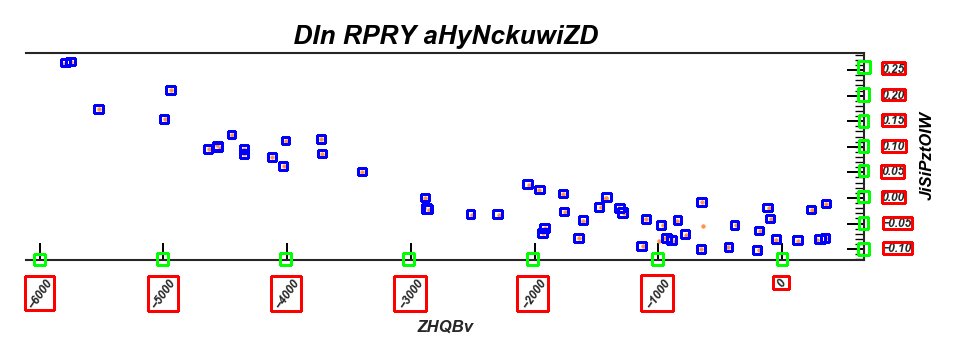}
  \end{minipage}
  \\
  \begin{minipage}[b]{0.48\textwidth}
	\includegraphics[width=\linewidth]{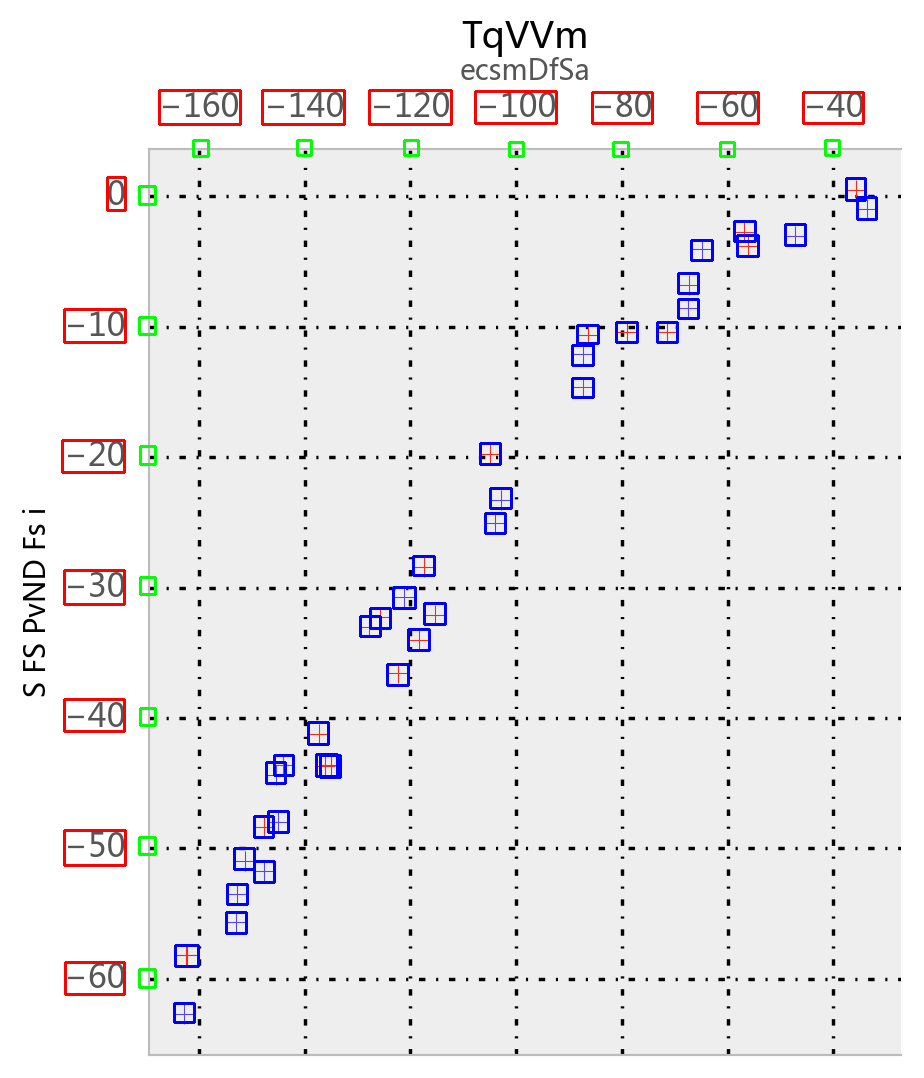}
  \end{minipage}
    \hfill
    \begin{minipage}[b]{0.48\textwidth}
	\includegraphics[width=\linewidth]{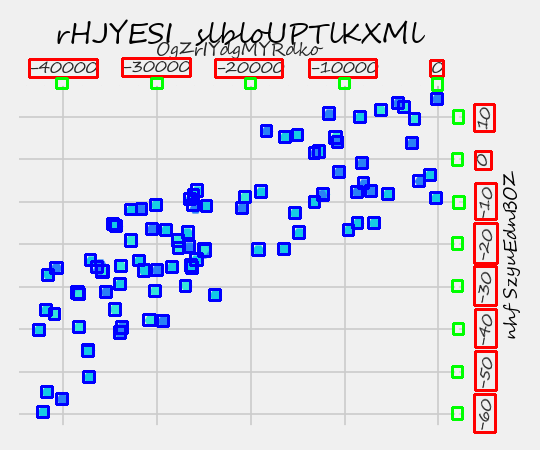}
  \end{minipage}
  \\
  \begin{minipage}[b]{0.48\textwidth}
	\includegraphics[width=0.7\linewidth]{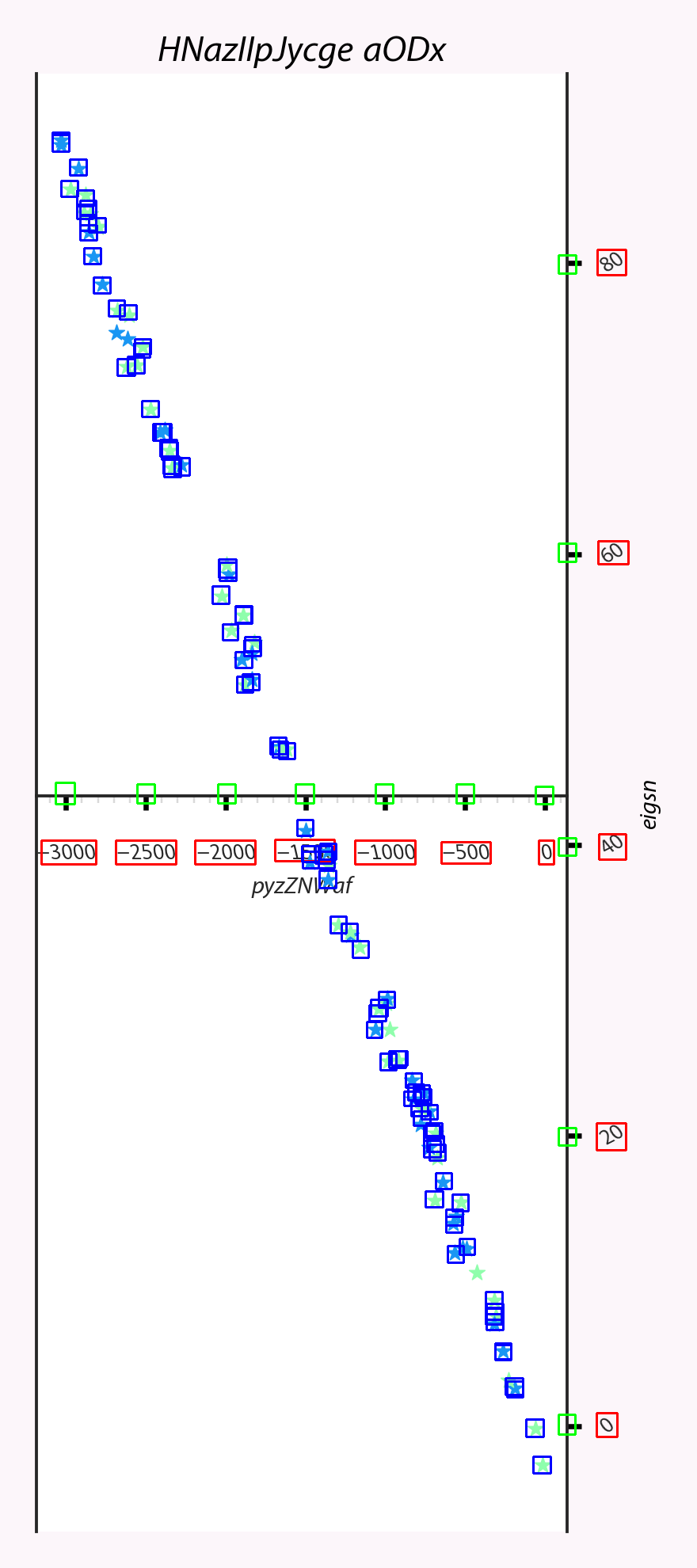}
  \end{minipage}
  \hfill
  \begin{minipage}[b]{0.48\textwidth}
	\includegraphics[width=\linewidth]{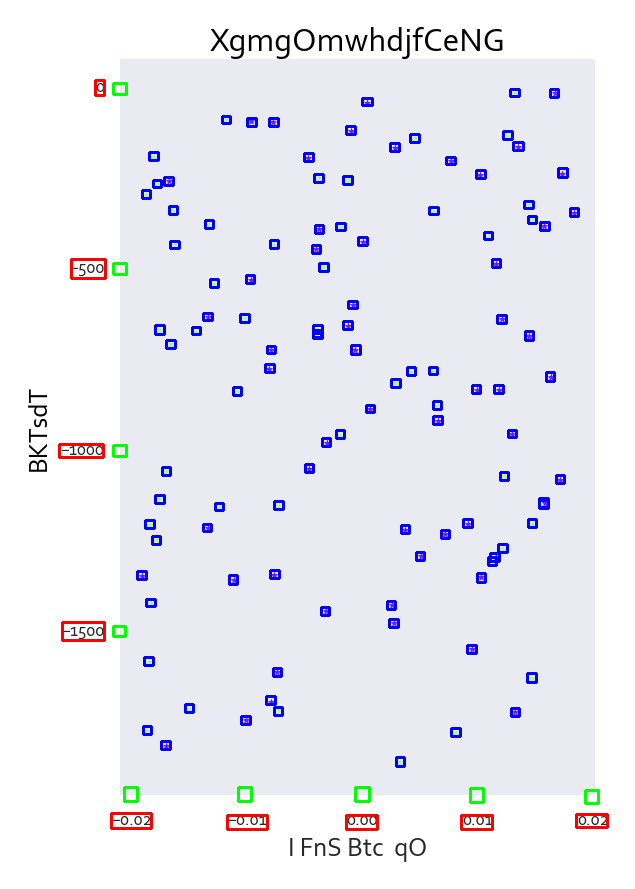}
  \end{minipage}
  \caption{Successful examples from the procedurally generated test set.}
\label{fig:suc_gen}  
\end{figure}

\begin{figure}[t]
  \centering
  \subfigure[\url{https://www.harrisgeospatial.com/docs/html/images/SedData_Scatterplot.png}]{\includegraphics[width=0.43\linewidth]{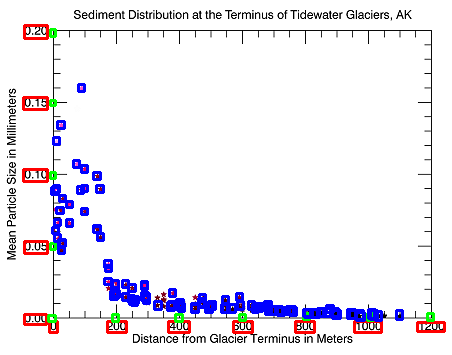}}
  \subfigure[\url{http://d2r5da613aq50s.cloudfront.net/wp-content/uploads/460860.image0.jpg}]{\includegraphics[width=0.43\linewidth]{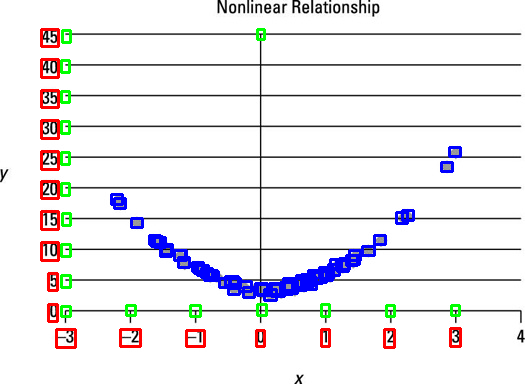}}
  \subfigure[\url{http://www.statisticshowto.com/wp-content/uploads/2013/08/spss-scatter-plot-4.png}]{\includegraphics[width=0.43\linewidth]{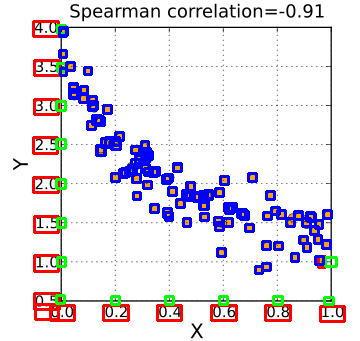}}
  \subfigure[\url{http://docs.enthought.com/chaco/_images/scatter_plot.png}]{\includegraphics[width=0.43\linewidth]{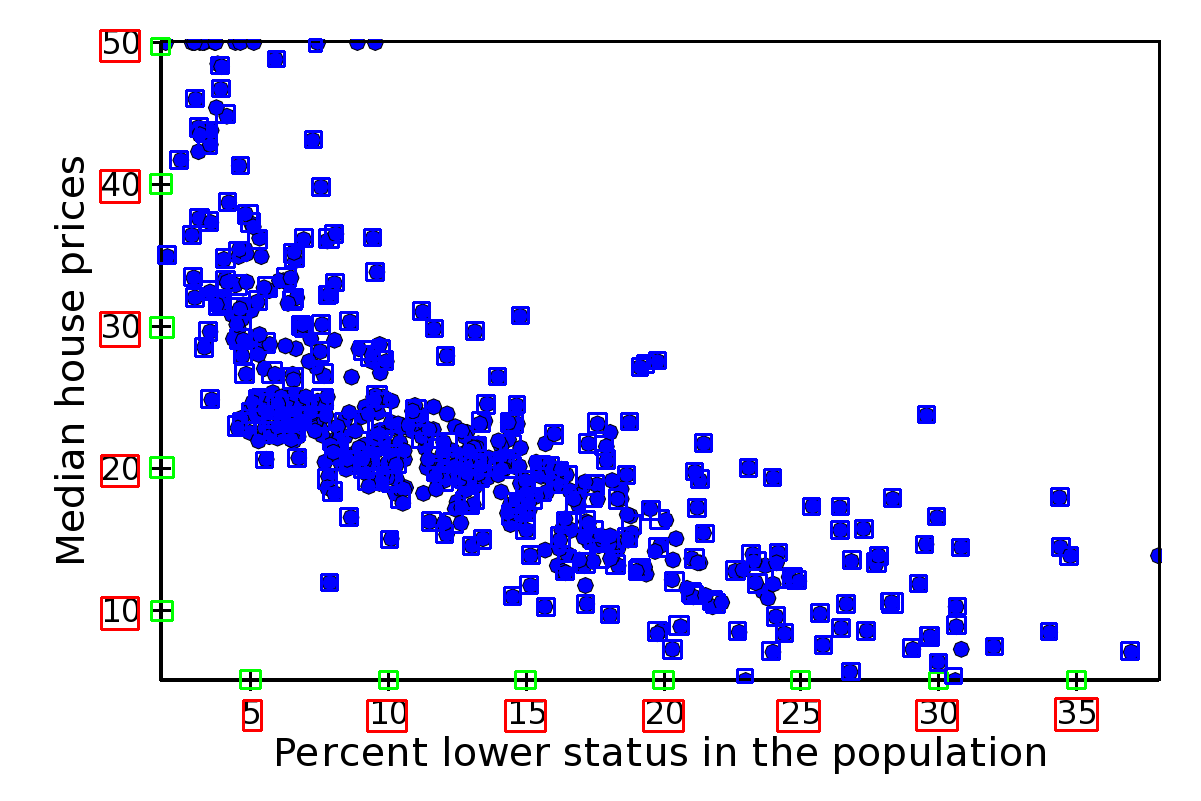}}
  \subfigure[\url{https://plot.ly/~RPlotBot/4326/styled-scatter.png}]{\includegraphics[width=0.43\linewidth]{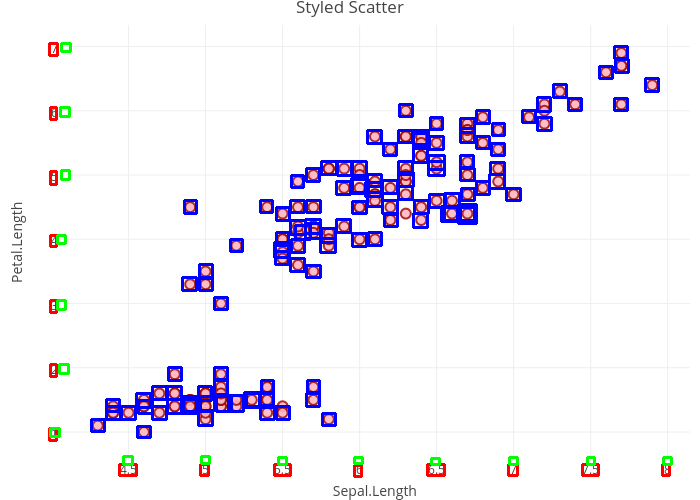}}
  \subfigure[\url{https://statsmethods.files.wordpress.com/2013/05/scatter-plot-2.png}]{\includegraphics[width=0.43\linewidth]{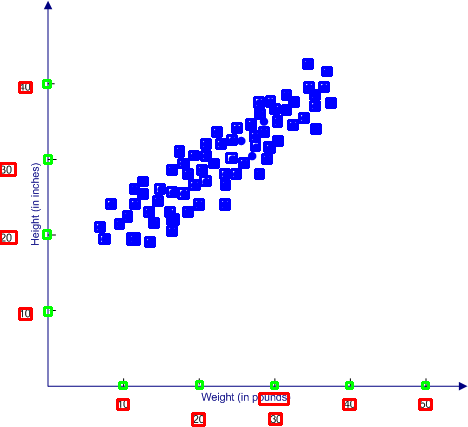}}
  \caption{Successful examples from the web test set.}
\label{fig:suc_web}  
\end{figure}

\begin{figure}[t]
\centering     
\subfigure[]{\label{fig:fail_gen_a}\includegraphics[width=0.40\linewidth]{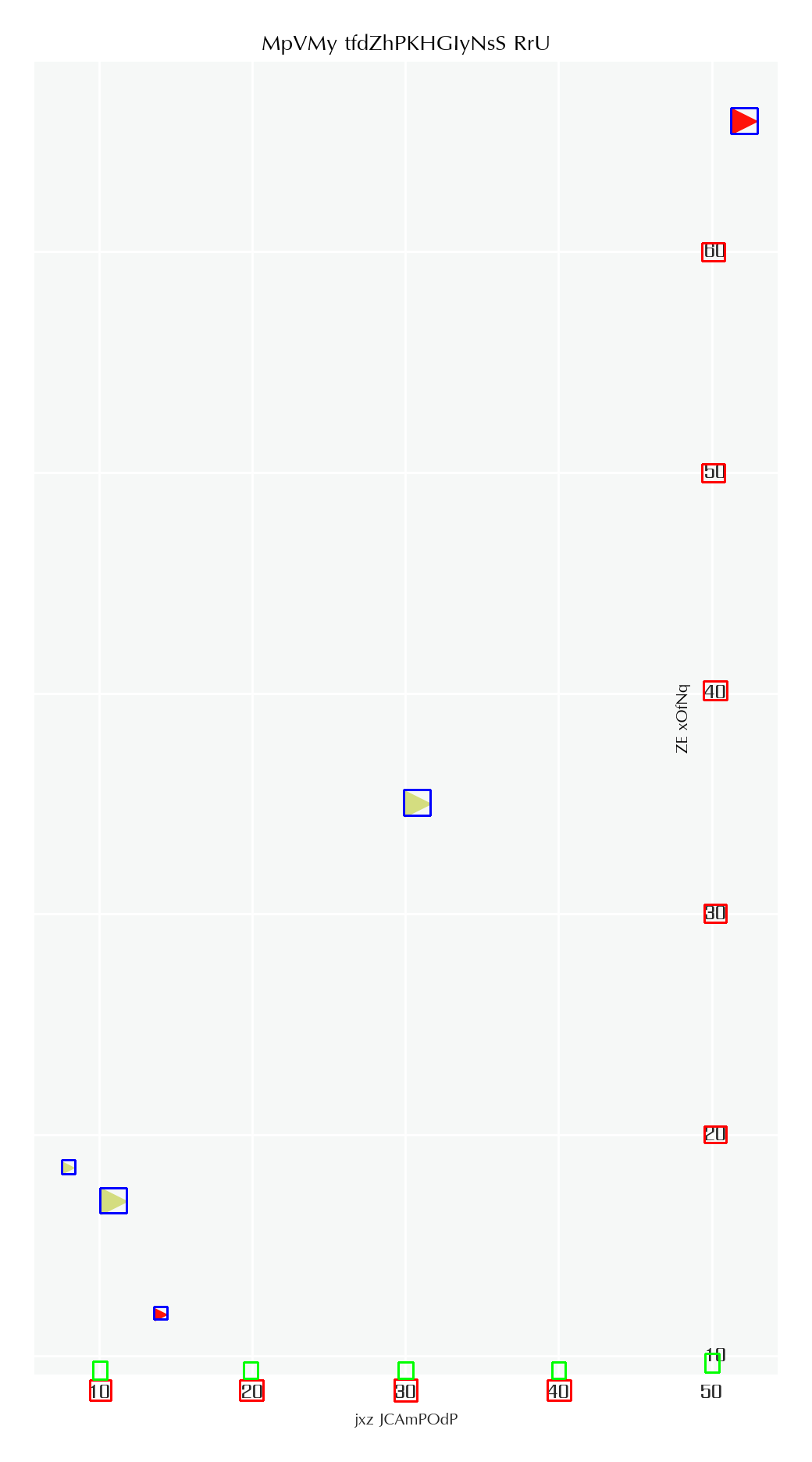}}
\subfigure[]{\label{fig:fail_gen_b}\includegraphics[width=0.40\linewidth]{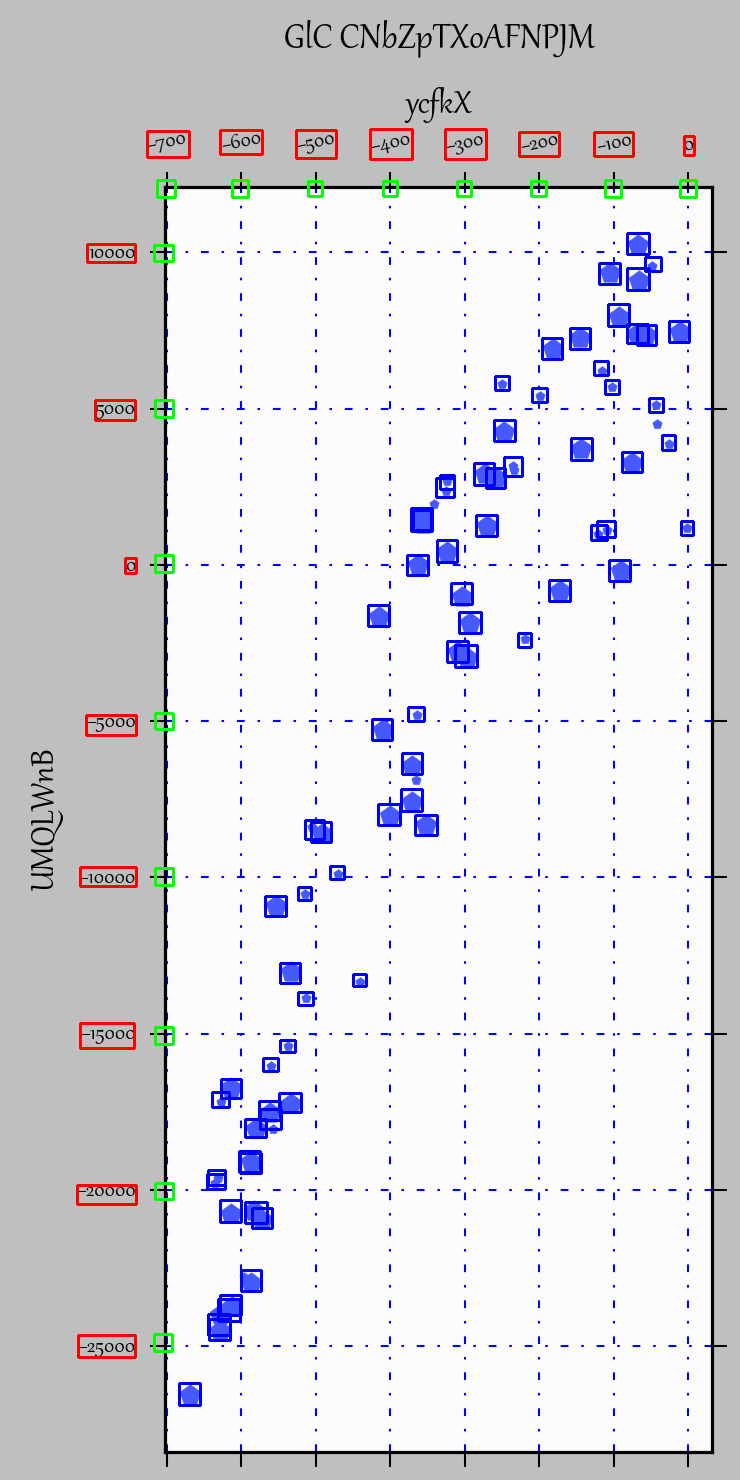}}
\caption{Unsuccessful examples from the procedurally generated test set.}
\label{fig:fail_gen}
\end{figure}

\begin{figure}[t]
\centering     
\subfigure[\url{http://www.wekaleamstudios.co.uk/wp-content/uploads/2010/04/scatterplot-base.jpeg}]{\label{fig:fail_web_a}\includegraphics[width=0.40\linewidth]{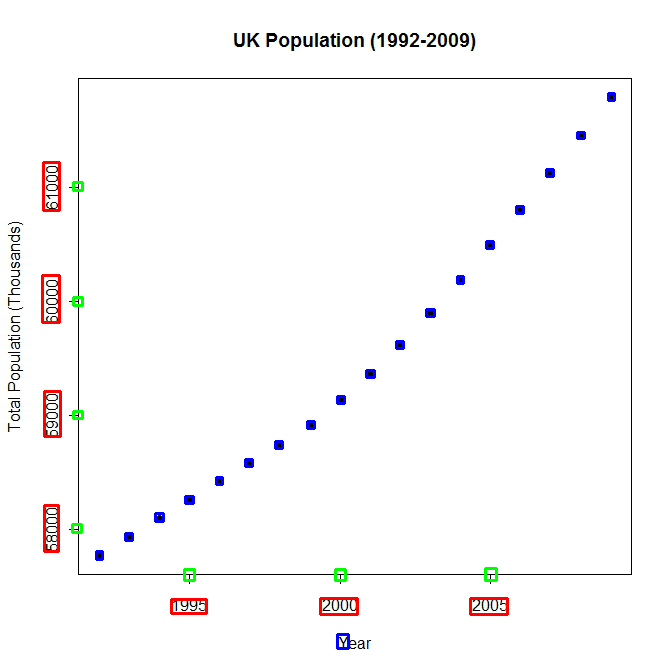}}
\subfigure[\url{https://www.ncl.ucar.edu/Applications/Images/scatter_2_1_lg.png}]{\label{fig:fail_web_c}\includegraphics[width=0.40\linewidth]{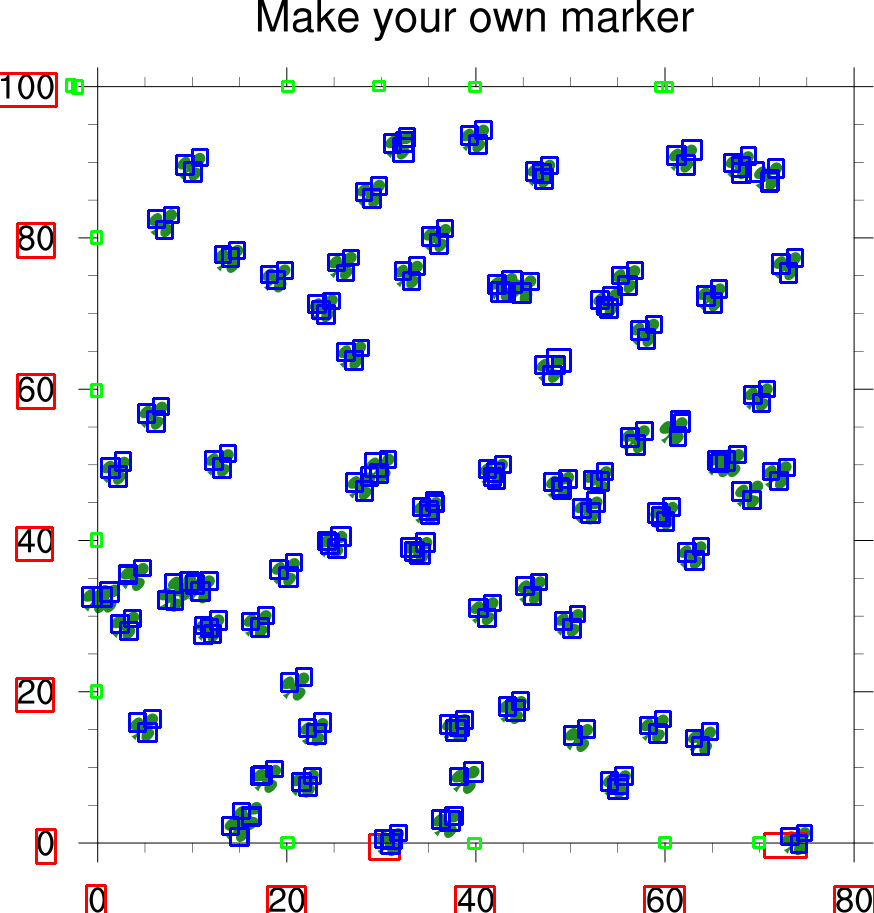}}
\caption{Unsuccessful examples from the web test set.}
\label{fig:fail_web}
\end{figure}

Let us now comment on results for individual plots.  First, Fig. \ref{fig:suc_gen} shows six successful data extraction plots from the procedurally generated test set, while Fig. \ref{fig:suc_web} displays six successful data extraction plots from the web test set.  Note that these plots cover a wide range of features occasionally seen in scatter plots.  Second, Fig. \ref{fig:fail_gen} displays two unsuccessful data extraction plots from the procedurally generated test set, while Fig. \ref{fig:fail_web} displays two unsuccessful data extraction plots from the web test set.  By examining the picture with the predicted bounding boxes and the result of the OCR we can see which part of the system failed on the unsuccessful examples.  On Fig. \ref{fig:fail_gen_a} the tick detection failed on the Y-axis because it is ambiguous whether the ticks should be on the left or right of the tick values, and on Fig. \ref{fig:fail_gen_b} the OCR failed to pick up the minus sign on several tick values of the X-axis.  Similarly, on Fig. \ref{fig:fail_web_a} our rotation fixing procedure fails on the 90$^o$-rotated Y tick values, and on Fig. \ref{fig:fail_web_c} the shamrock marker is detected as three separate points by the object detection.  Those errors represent well the kind of failure that our method can encounter.  

\section{Conclusion}\label{conclusion}

In this paper we introduced Scatteract, a method used to automatically extract data from scatter plots.  Arguably the most important innovation in Scatteract is the deep learning object detection models used to locate points, tick marks, and tick values, combined with the ability to train these models on procedurally generated scatter plots, for which the ground truth is known.  The other main contribution of Scatteract is the pipeline, from the OCR preprocessing to the RANSAC regression, for finding the affine transformation between pixel coordinates and chart coordinates.  There seems to be relatively little prior work on this, and we have shown our method to be more robust than the other methods we are aware of in the literature.  The end-to-end system is able to successfully extract data from 89\% of procedurally generated scatter plots and 78\% of scatter plots from the web. \\

Extending our models to handle new chart aesthetics is straightforward if we can generate training data with the new aesthetics.  While the performance on the web test set is promising, it is somewhat worse than performance on the procedurally generated test set.  We leave to future work an investigation of how to improve Scatteract's generalization ability.  We will also pursue a deeper extraction of information from scatter plots, including point categories, legends, axis titles, nonlinear scales, scales with units, and non-numeric scales, such as dates.  The ultimate goal is to extend the method to other chart types as well, so that an end-to-end system would detect charts, classify them (line chart, bar chart, pie chart, scatter plot, etc.), and then use an appropriate pipeline to extract the data from the chart.

\bibliography{plot_extraction}
\bibliographystyle{splncs03}

\end{document}